\documentclass{article}
\usepackage{spconf,amsmath,graphicx}
 \usepackage{amssymb}
 \usepackage{pifont}
\usepackage{subfigure} 
\usepackage{float}
\usepackage{stfloats}
\usepackage{booktabs}
\usepackage{setspace}
\usepackage{multirow}
\usepackage{cite}
\usepackage{enumerate} 
\title{Refined Pseudo labeling for Source-free Domain Adaptive Object Detection}
\name
{
	Siqi~Zhang\textsuperscript{1,2}\quad Lu~Zhang\textsuperscript{1} \quad Zhiyong~Liu\textsuperscript{1,2,3}\sthanks{Corresponding author. This work was supported in part by the National Key Research and Development Plan of China under Grant 2020AAA0108902, the Strategic Priority Research Program of Chinese Academy of Science under Grant XDB32050100 and the NSFC under Grant 62206288.} % <-this % stops a space
}
\address{\textsuperscript{1} Institute of Automation, Chinese Academy of Sciences, Beijing, China \\
\textsuperscript{2}School of Artificial Intelligence, University of Chinese Academy of Sciences, Beijing, China \\
\textsuperscript{3}Nanjing Artificial Intelligence Research of IA,  Jiangsu, China }
\begin{document}
%\ninept
%
\maketitle
\begin{abstract}
Domain adaptive object detection (DAOD) assumes that both labeled source data and unlabeled target data are available for training, but this assumption does not always hold in real-world scenarios. Thus, source-free DAOD is proposed to adapt the source-trained detectors to target domains with only unlabeled target data. Existing source-free DAOD methods typically utilize pseudo labeling, where the performance heavily relies on the selection of confidence threshold. However, most prior works adopt a single fixed threshold for all classes to generate pseudo labels, which ignore the imbalanced class distribution, resulting in biased pseudo labels. In this work, we propose a refined pseudo labeling framework for source-free DAOD. First, to generate unbiased pseudo labels, we present a category-aware adaptive threshold estimation module, which adaptively provides the appropriate threshold for each category. Second, to alleviate incorrect box regression, a localization-aware pseudo label assignment strategy is introduced to divide labels into certain and uncertain ones and optimize them separately. Finally, extensive experiments on four adaptation tasks demonstrate the effectiveness of our method.
\end{abstract}
\section{Introduction}
\label{sec:intro}
Convolution neural networks have achieved remarkable success for object detection when trained on a large amount of labeled data \cite{ren2015faster,tian2019fcos,bochkovskiy2020yolov4}. However, well-trained networks often suffer from performance degradation when deployed from the source domain to the target domain, due to domain shifts such as light variations, weather changes, or style differences. To tackle this problem, domain adaptive object detection (DAOD)~\cite{chen2018domain,saito2019strong,xu2020exploring,su2020adapting,he2020domain,Chen_2020_CVPR,deng2021unbiased} has been widely explored to mitigate the domain shift by transferring knowledge from labeled source data to unlabeled target data. Existing DAOD methods~\cite{chen2018domain,saito2019strong,xu2020exploring,su2020adapting,he2020domain,deng2021unbiased} typically assume that the source data is available during training phase. However, due to inefficient data transmission, data storage and privacy concerns, the access \begin{figure}[t]
\centering
\includegraphics[width=0.48\textwidth]{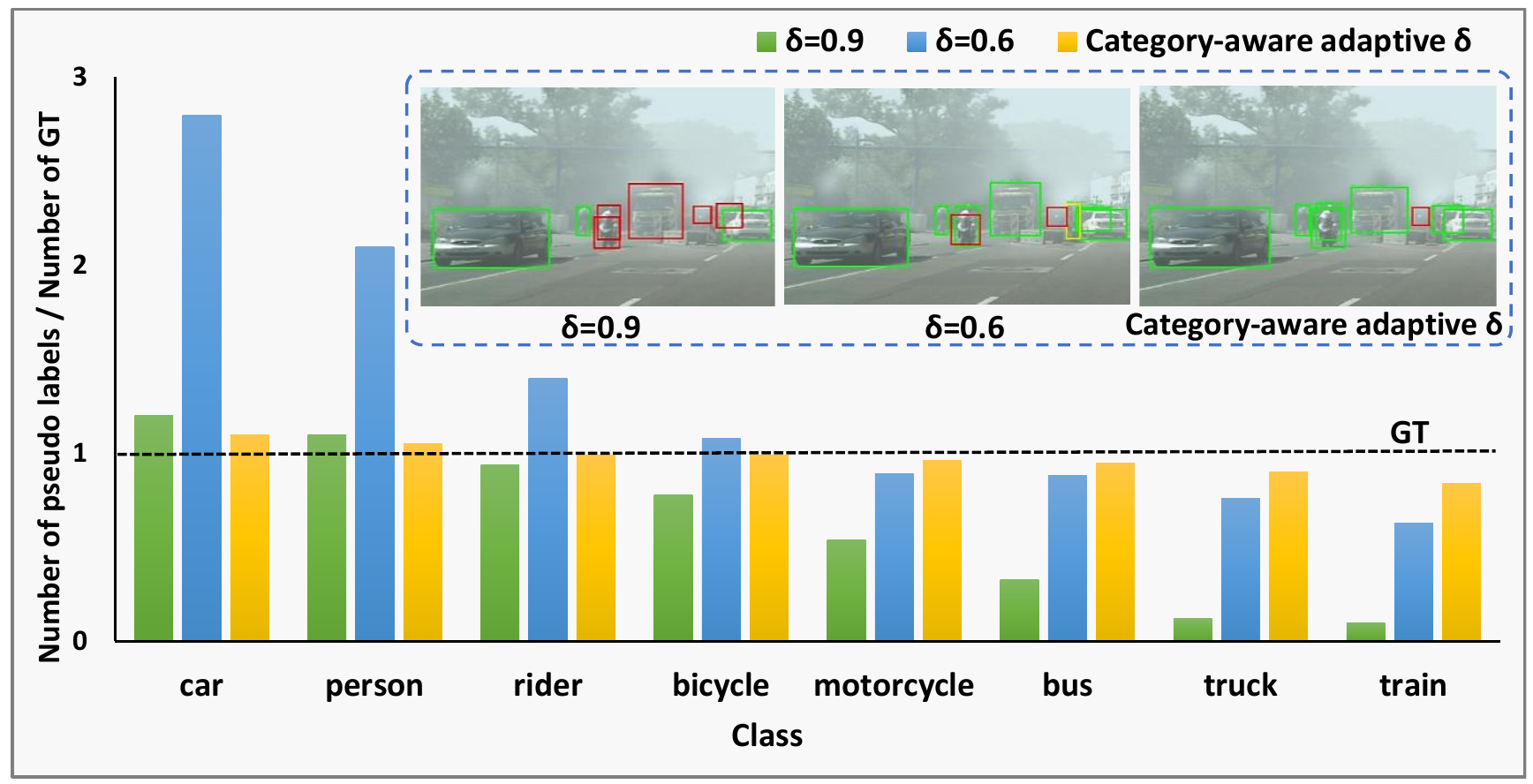}
\vspace{-2em}
\caption{There is a bias between the true labels and the pseudo labels generated by a fixed confidence threshold $\delta$ due to the class-imbalanced data distribution. when $\delta$ is set to 0.6, the threshold will filter out many high-quality labels, leading to false positives. Conversely, when $\delta$ is set to a high value of 0.9, the number of false negatives (red rectangles) will increase.}
\label{fig1}
\vspace{-1.6em}
\end{figure}to source data is not always practical in real-world scenarios. 

Hence, source-free DAOD ~\cite{li2021free,huang2021model,li2022source}, where only a model pre-trained on the source domain and unlabeled target data are provided for adaptation, has drawn more attention. The major challenge of source-free DAOD is the absence of labeled source data, and traditional DAOD techniques (adversarial learning~\cite{chen2018domain,saito2019strong,xu2020exploring,su2020adapting} and image translation~\cite{Chen_2020_CVPR,deng2021unbiased}) are not applicable. To overcome this challenge, pseudo labeling~\cite{li2021free,tarvainen2017mean} is introduced for source-free DAOD, where pseudo labels for target samples are selected through filtering pseudo boxes generated by the source pre-trained detector via a hand-crafted category confidence threshold $\delta$. In this paradigm, the detection performance is heavily related to the quality of pseudo labels, which is sensitive to the selection of the threshold. For high-quality pseudo label generation, Li \textit{et al}.~\cite{li2021free} proposes a metric named self-entropy descent to search the appropriate threshold. LODS ~\cite{li2022source} overlooks the target domain style for reliable pseudo labels. Despite these efforts, there are still two obstacles hindering the improvement of detection performance.

The first obstacle is class-imbalanced data distribution. As shown in Fig.\ref{fig1}, there is a bias between the true labels and the pseudo labels generated by the fixed confidence threshold $\delta$ during self-training~\cite{wang2022category}. 
\begin{figure*}[ht]
\centering
\includegraphics[width=0.75\textwidth]{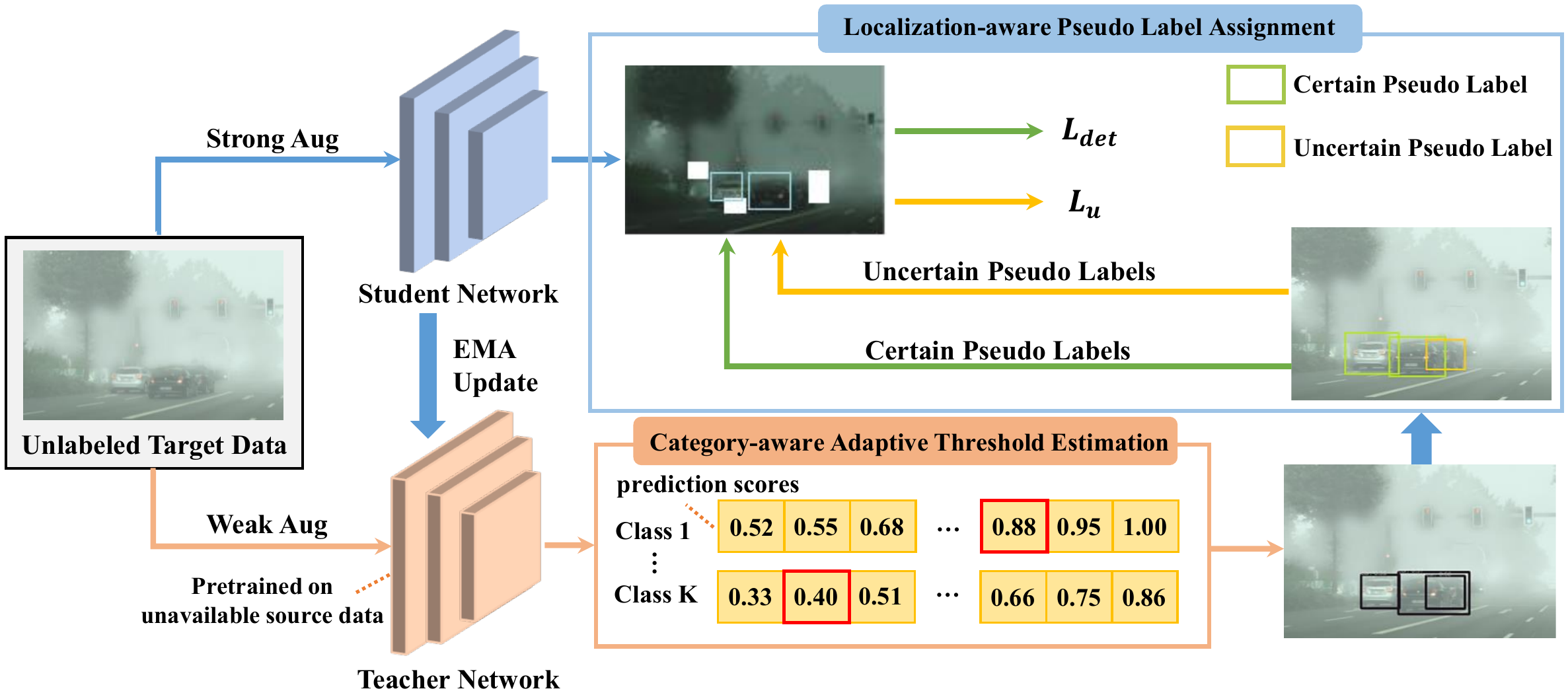}
\vspace{-0.5em}
\caption{Overview of the proposed framework. The pseudo labels of unlabeled target data are generated by the teacher network according to the category-aware adaptive threshold. Then, we introduce localization-aware pseudo label assignment to split pseudo labels into uncertain and certain ones and optimize them with different strategies.}
\label{fig2}
\vspace{-1em}
\end{figure*}The number of pseudo labels in some classes (\textit{e.g.}, car and person) is much higher than the number of ground-truth labels, while much lower in others(\textit{e.g.}, truck and train). Moreover, when $\delta$ is set to a high value of 0.9, the threshold will filter out many high-quality labels, leading to false negatives. Conversely, when $\delta$ is set to 0.6, the number of false positives will increase. As a result, it is hard to find a perfect global threshold $\delta$ to guarantee the quality of pseudo labels. Another obstacle is the inaccuracy localization of pseudo labels. As for object detection, the pseudo label consists of both category labels and bounding boxes. Since only category confidence is considered to filter out low-quality pseudo boxes~\cite{li2021free,tarvainen2017mean}, localization accuracy cannot be guaranteed.

To overcome these obstacles, in this paper, we propose a refined pseudo labeling (RPL) method for source-free DAOD based on self-training in a mean-teacher framework. To deal with category-biased pseudo labeling, we propose a category-aware adaptive threshold estimation (CATE) module to produce unbiased pseudo labels for target samples. The threshold is set according to the distributions of each category in the foreground predictions of the teacher network, which is category-specific. Furthermore, as the teacher network is gradually optimized during training, we refresh the thresholds. To address the inaccurate localization, we introduce a localization-aware pseudo label assignment (LPLA) strategy to improve the localization quality of pseudo labels. Specially, according to box regression variance, the pseudo labels are divided into uncertain and certain ones. Certain ones are exploited as supervised information for student network optimization, while uncertain ones are utilized in a soft learning manner. In detail, the regression loss for uncertain ones is removed to avoid incorrect box regression, and we adopt teacher-generated proposals corresponding to uncertain pseudo-labels to obtain matching features in the student network to construct the loss. We conduct extensive experiments on four different domain adaptation scenarios and the results demonstrate the superiority of our proposed method over state-of-the-arts.

 The main contributions of this paper are listed as follows: (1) We explore the obstacles in source-free DAOD: category-biased pseudo label generation and inaccurate localization and propose a simple yet effective RPL framework. (2) In the RPL framework, we present a category-aware adaptive threshold estimation module to mitigate the biased pseudo label generation and a localization-aware pseudo label assignment strategy to alleviate poor box regression. (3) Extensive experiments on four adaptation tasks demonstrate the effectiveness of our method.

 \vspace{-0.5em}
\section{Method}
\label{sec:method}
 \vspace{-1em}
\subsection{Preliminaries}
\textbf{Task Definition.} In the source-free DAOD setting, only a source domain pre-trained model $\theta$ and the unlabeled target dataset $\mathcal{D}_{t}=\left\{x_{t}^{i}\right\}_{i=1}^{n_{t}}$ are given during adaptation, and the source dataset is unavailable.

\noindent
\textbf{Mean Teacher Framework.} Previous work~\cite{french2017self} has found it is applicable to simulate unsupervised domain adaptation as semi-supervised learning. Thus, we build our approach based on the mean-teacher framework~\cite{tarvainen2017mean}, which is an effective technique in semi-supervised learning. Mean-teacher framework~\cite{tarvainen2017mean} contains a teacher network (pre-trained on the source domain)  $\theta_t$ to generate pseudo labels and a student network $\theta_s$ (initialized as $\theta_t$) to update the teacher network via Exponential Moving Average (EMA). In the training phase, pseudo labels are generated on the weakly augmented unlabeled samples to serve as supervision for the corresponding strong augmentation. The training objective can be summarized as:
\vspace{-0.6em}
\begin{equation}
\mathcal{L}_{sl}=\sum_i \mathcal{L}_{cls}\left(x_t^i, \tilde{y}_t^i\right)+\mathcal{L}_{reg}\left(x_t^i, \tilde{y}_t^i\right),
\vspace{-0.6em}
\label{1}
\end{equation}
where $\tilde{y}_t$ is pseudo labels generated by the teacher, $\mathcal{L}_{cls}$ and $\mathcal{L}_{reg}$ denote classification and box regression loss respectively. Furthermore, the teacher is gradually updated via EMA of student weight.

Due to the domain shift, the pseudo labels generated by the source-trained teacher are noisy, which harms detection performance on the target domain. To improve the quality of pseudo labels, we propose refined pseudo labeling (RPL), including category-aware adaptive threshold estimation and localization-aware pseudo label assignment.
 \vspace{-0.5em}
 \begin{table}[t]
\centering
\caption{Detection results (\%) on  Cityscapes to Foggy Cityscapes. %'No DA' indicates the model is only trained with the source images and directly tested on the target images without any domain adaptation. 
The best results are in \textbf{bold}.}
\resizebox{\linewidth}{!}{
\begin{tabular}{l|c|cccccccc|c}
\toprule[1.5pt]
Methods     &Source & person & rider & car & truck & bus & train & motor & bicycle & mAP \\ \hline
No DA &\checkmark&25.8  &33.3 &35.2 & 13.0 &26.4& 9.1 &19.0&32.3&24.3\\\hline
DA~\cite{chen2018domain}&\checkmark&25.0 &31.0 & 40.5& 22.1 & 35.3& 20.2 &20.0 &  27.1 &  27.6   \\
SWDA~\cite{saito2019strong}&\checkmark& 29.9 &42.3  &43.5& 24.5& 36.2& 32.6 &30.0&  35.3 &34.3 \\
CDN~\cite{su2020adapting} &\checkmark& 35.8& 45.7 &50.9& 30.1& 42.5& 29.8& 30.8& 36.5& 36.6\\
HTCN~\cite{Chen_2020_CVPR}&\checkmark&33.2& 47.5 &47.9& 31.6 &47.4& 40.9 &32.3& 37.1& 39.8\\
ATF~\cite{he2020domain} &\checkmark& 34.6 &47.0&  50.0 & 23.7&  43.3  &38.7  &33.4  & 38.8 &38.7\\ 
UMT~\cite{deng2021unbiased}&\checkmark& 33.0 &46.7  &48.6& 34.1& 56.5&46.8&30.4&37.3& 41.7\\ \hline
SFOD~\cite{li2021free}      &\ding{53}&   21.7& 44.0& 40.4& 32.2& 11.8 &25.3 &34.5& 34.3 &30.6  \\
SFOD-Mosaic~\cite{li2021free}      &\ding{53}&25.5& 44.5& 40.7& \textbf{33.2}& 22.2& \textbf{28.4}& 34.1& 39.0& 33.5\\
HCL~\cite{huang2021model}      &\ding{53}&26.9& 46.0& 41.3& 33.0& 25.0& 28.1& \textbf{35.9}& 40.7& 34.6\\
Mean-Teacher~\cite{tarvainen2017mean}&\ding{53}& 33.9 &43.0& 45.0 &29.2 &37.2& 25.1& 25.6& 38.2 &34.3\\
LODS~\cite{li2022source}      &\ding{53}&   34.0    &  45.7   &  48.8  & 27.3    & 39.7   & 19.6  & 33.2  & 37.8      &  35.8 \\ \hline
RPL       &\ding{53}&   \textbf{37.9}    &   \textbf{52.7}   &   \textbf{52.0}  & 29.9    & \textbf{46.7}   & 21.7   & 34.5  & \textbf{46.5}     &  \textbf{40.2}  \\ 
\bottomrule[1.5pt]
\end{tabular}}
\label{tab:city}
\vspace{-1em}
\end{table}
\subsection{Category-aware Adaptive Threshold Estimation}
Real-world datasets usually exhibit long-tailed distributions. Due to the imbalanced class distribution, it would lead to pseudo labels biased to the majority classes if all categories adopt a single fixed confidence threshold. To alleviate this problem, we introduce a category-aware adaptive threshold estimation (CATE) module. CATE could select high thresholds for the majority categories and assign low thresholds to the minor categories adaptively. Taking the urban road scene as an example, the confidence thresholds of the car and person for generating pseudo labels are higher than the truck and train. Specifically, giving a subset of unlabeled data to the teacher network, the confidence threshold $\delta_{i}$ for the $i$-th category to select pseudo boxes is estimated as:
\vspace{-0.5em}
\begin{equation}
\delta_{i}=L_i^{sort}\left[n_i \cdot P_i\right], P_i=\frac{n_i}{n_f}
\vspace{-0.5em}
\end{equation}
Where $L_i^{sort}$ is an ascending list sorted by prediction scores of the $i$-th category, $n_f$ denotes the box number of foreground, $n_i$ is the box number of the $i$-th category and $P_i$ denotes the proportion of the $i$-th category in the foreground.

As the teacher network is gradually optimized during training, we refresh the thresholds every 500 iterations to adapt to the current teacher network dynamically. The thresholds we design are adaptively updated and category-aware, which benefits the generation of unbiased pseudo labels.

\begin{table}[t]
\centering
\caption{Detection results (\%) on  Sim10K to Cityscapes and KITTI to Cityscapes.}
\resizebox{0.66\linewidth}{!}{
\begin{tabular}{l|c|c|c}
\toprule[1.5pt]
\multicolumn{1}{l|}{\multirow{2}{*}{Methods}} & \multicolumn{1}{l|}{\multirow{2}{*}{Source}} & \multicolumn{1}{c|}{SIM10K $\rightarrow$ City} & KITTI $\rightarrow$ City \\ \cline{3-4} 
\multicolumn{1}{l|}{}                        & \multicolumn{1}{l|}{}                        & \multicolumn{1}{c|}{AP of car}   & AP of car  \\\hline
No DA &\checkmark&35.8  &33.3 \\\hline
DA~\cite{chen2018domain}&\checkmark& 38.9 &38.5   \\
SWDA~\cite{saito2019strong}&\checkmark& 40.1 &37.9 \\
CDN~\cite{su2020adapting} &\checkmark& 49.3 &44.9 \\
HTCN~\cite{Chen_2020_CVPR}&\checkmark&42.5&42.8\\
ATF~\cite{he2020domain} &\checkmark&42.8  &42.1\\
UMT~\cite{deng2021unbiased}&\checkmark& 43.1 &43.9 \\ \hline
SFOD~\cite{li2021free}      &\ding{53}& 42.3& 43.6 \\
SFOD-Mosaic~\cite{li2021free}      &\ding{53}&42.9& 44.6\\
Mean-Teacher~\cite{tarvainen2017mean}   &\ding{53}&   40.5   &  41.7   \\ 
LODS~\cite{li2022source}      &\ding{53}&   45.7    &   43.9 \\ \hline
RPL        &\ding{53}&   \textbf{50.1}    &   \textbf{47.8}    \\ 
\bottomrule[1.5pt]
\end{tabular}}
\label{tab:sim}
\vspace{-1em}
\end{table}
 \vspace{-0.5em}
\subsection{Localization-aware Pseudo label Assignment}
Since there is not a strong positive correlation between the localization accuracy and the foreground score of the box candidates~\cite{xu2021end}, the boxes with high foreground scores could provide inaccurate localization. This indicates that simply considering category scores can not guarantee the quality of pseudo boxes. To avoid incorrect box regression, we present a localization-aware pseudo label assignment (LPLA) strategy to divide pseudo labels into certain and uncertain ones. Based on the observation that the box regression variance could measure the localization accuracy~\cite{xu2021end}, we take mean Intersection over Union (IoU) $M_{IoU}$ as the basis of division. A group of neighboring boxes is suppressed to a single bounding box after Non-Maximum Suppression, and mean IoU $M_{IoU}$ is the average IoU score between the pseudo box and corresponding suppressed ones. 

Pseudo labels with $M_{IoU}>\beta$ are divided into certain ones, which are used as supervised labels for student network training. The supervised loss is defined as:
\vspace{-0.5em}
\begin{equation}
\mathcal{L}_{det}=\sum_i \mathcal{L}_{cls}\left(x_t^i, \tilde{y}_{tc}^i\right)+\mathcal{L}_{reg}\left(x_t^i, \tilde{y}_{tc}^i\right),
\vspace{-0.5em}
\end{equation}
where $\tilde{y}_{tc}$ denotes the certain pseudo labels. For the uncertain ones, we remove the box regression loss for them to avoid inaccuracy box regression. To construct the loss $L_{u}$ for uncertain labels, we adopt teacher-generated proposals corresponding to uncertain pseudo labels to obtain matching features in the student, and the loss $L_{u}$ for uncertain labels is defined as:
\vspace{-0.5em}
\begin{equation}
\mathcal{L}_{u}=\sum_{i=1}^{n_p}\sum_{j=1}^{C}-p_{i,j}^t\log p_{i,j}^s,
\vspace{-0.5em}
\end{equation}
where $C$ is the class number, $p_{i,j}^t$ and $p_{i,j}^s$ is the probability of the $j$-th category in $i$-th proposal from the teacher and student network and $n_p$ is the number of the proposals corresponding to uncertain pseudo labels. Finally, we reformulate the training objective in Eq.\ref{1} as: 
\vspace{-0.5em}
\begin{equation}
\mathcal{L}_{sl}=\mathcal{L}_{det} + \mathcal{L}_{u}.
\vspace{-0.5em}
\end{equation}
 The overall framework updates as:
 \vspace{-0.5em}
\begin{equation}
\theta_s \leftarrow \theta_s+\gamma \frac{\partial\left(\mathcal{L}_{sl}\right)}{\partial \theta_s},
\vspace{-0.5em}
\end{equation}
\begin{equation}
\theta_t \leftarrow \alpha \theta_t+(1-\alpha) \theta_s,
\end{equation}
where $\gamma$ is the student learning rate and $\alpha$ is the EMA coefficient. 
\section{Experiment}
\label{sec:experiment}
\vspace{-0.5em}
\subsection{Experimental Setup}
\textbf{Datasets.} We utilize six public datasets in our experiments, including four different domain adaptation scenarios: weather adaptation (Cityscapes~\cite{cordts2016cityscapes} to Foggy Cityscapes~\cite{sakaridis2018semantic}), synthesis to real world adaptation (Sim10k~\cite{johnson2017driving} to Cityscapes), camera configuration adaptation (KITTI~\cite{geiger2013vision} to Cityscapes) and real to art adaptation (Pascal~\cite{everingham2012pascal} to Clipart~\cite{inoue2018cross}).
\begin{table*}[t]
\centering
\caption{Detection results (\%) on  Pascal to Clipart.}
\resizebox{\linewidth}{!}{
\begin{tabular}{l|c|cccccccccccccccccccc|c}
\toprule[1.5pt]
Methods     &Source & aero& bcycle& bird &boat& bottle &bus& car &cat& chair& cow& table& dog& hrs& bike& prsn& plnt& sheep& sofa& train& tv& mAP \\ \hline
No DA &\checkmark&35.4& 52.8& 24.9 &23.4 &22.0& 48.5& 32.7& 10.8 &30.9 &11.0 &13.3& 10.3 &25.1 &45.3& 48.6 &37.8 &10.5& 7.8& 22.9& 29.8& 28.5\\\hline
DA~\cite{chen2018domain}&\checkmark&15.0& 34.6& 12.4& 11.9& 19.8 &21.1& 23.2 &3.1& 22.1 &26.3& 10.6& 10.0& 19.6& 39.4& 34.6& 29.3 &1.0& 17.1 &19.7& 24.8 &19.8   \\
SWDA~\cite{saito2019strong}&\checkmark& 26.2& 48.5& 32.6& 33.7& 38.5& 54.3& 37.1& 18.6& 34.8 &58.3 &17.0 &12.5 &33.8 &65.5 &61.6 &52.0 &9.3 &24.9 &54.1 &49.1 &38.1 \\
CRDA~\cite{xu2020exploring} &\checkmark&28.7& 55.3& 31.8& 26.0& 40.1& 63.6& 36.6& 9.4& 38.7& 49.3& 17.6& 14.1& 33.3& 74.3& 61.3& 46.3& 22.3& 24.3& 49.1& 44.3& 38.3\\
HTCN~\cite{Chen_2020_CVPR}&\checkmark&
33.6& 58.9& 34.0& 23.4& 45.6& 57.0& 39.8&12.0& 39.7& 51.3& 21.1& 20.1& 39.1& 72.8& 63.0& 43.1& 19.3& 30.1& 50.2& 51.8& 40.3\\
ATF~\cite{he2020domain} &\checkmark& 41.9 &67.0 &27.4 &36.4 &41.0& 48.5& 42.0& 13.1& 39.2& 75.1& 33.4 &7.9 &41.2 &56.2 &61.4& 50.6& 42.0 &25.0& 53.1 &39.1 &42.1\\ 
UMT~\cite{deng2021unbiased}&\checkmark& 39.6 &59.1 &32.4& 35.0 &45.1 &61.9& 48.4& 7.5& 46.0& 67.6 &21.4& 29.5& 48.2& 75.9& 70.5& 56.7 &25.9& 28.9& 39.4& 43.6 &44.1\\ \hline
SFOD~\cite{li2021free}      &\ding{53}&  20.1& 51.5 &26.8 &23.0& 24.8& \textbf{64.1} &37.6& 10.3& 36.3& 20.0 &18.7& 13.5& 26.5& 49.1& 37.1& 32.1& 10.1& 17.6 &42.6 &30.0& 29.5  \\
Mean-Teacher~\cite{tarvainen2017mean}&\ding{53}& 30.9& 51.8& 27.2& 28.0& 31.4& 59.0& 34.2& 10.0& 35.1& 19.6& 15.8& 9.3 &\textbf{41.6}& 54.4& 52.6& 40.3& 22.7& 28.8& 37.8& 41.4& 33.6\\
LODS~\cite{li2022source} &\ding{53}& \textbf{43.1}& 61.4& \textbf{40.1}& \textbf{36.8}& 48.2& 45.8& 48.3& 20.4& 44.8&53.3& \textbf{32.5}& \textbf{26.1}& 40.6& \textbf{86.3}& \textbf{68.5}& 48.9& 25.4& \textbf{33.2}& 44.0& \textbf{56.5}& \textbf{45.2}\\
 \hline
RPL        &\ding{53}& 38.5 &\textbf{73.4} &34.4& 33.2 &\textbf{49.2}&62.3& \textbf{54.1}& \textbf{23.8}& \textbf{52.7}& \textbf{53.5} &29.9& 13.5& 32.4& 68.7& 61.7& \textbf{52.3} &\textbf{27.1}& 27.2& \textbf{53.7}& 37.7 &44.0 \\ 
\bottomrule[1.5pt]
\end{tabular}}
\label{tab:voc}
\vspace{-2em}
\end{table*}
\begin{table}[t]
\centering
\caption{Ablation study on four adaptation tasks.}
\resizebox{\linewidth}{!}{
\begin{tabular}{l|c|c|c|c}
\toprule[1.5pt]
Methods     &City to Foggy & Sim10k to City & KITTI to City & Pascal to  Clipart  \\ \hline
RPL         &\textbf{40.2} &  \textbf{50.1}    &  \textbf{47.8}   &   \textbf{44.0}   \\ 
RPL  w/o LPLA     &37.8 (-2.4) &   47.6  (-2.5)  &   45.6 (-2.2)   &  42.1 (-1.9)   \\ 
RPL  w/o CATE ($\delta$=0.7)  &36.8 (-3.4) &  47.0 (-3.1) &   44.5 (-3.3)  &   39.3 (-4.7) \\ 
RPL  w/o CATE ($\delta$=0.9)  & 32.3 (-7.9) &  42.7 (-7.4) & 43.3 (-4.5)    &   35.2 (-8.8) \\ 
\bottomrule[1.5pt]\end{tabular}}
\label{tab:ab}
\vspace{-1em}
\end{table}

\noindent
\textbf{Implementation Details.} For fair comparisons, we follow the same experimental setting as~\cite{saito2019strong,li2021free,li2022source}, and Faster R-CNN~\cite{ren2015faster} is used as the base detector. The source model is trained using SGD optimizer with a learning rate of 0.001 and momentum of 0.9. For the proposed framework, the EMA rate $\alpha$ is set to 0.99, $\beta$ is set to 0.85 and the student learning rate $\gamma$ is set to 0.001. We report the mean Average Precision (mAP) with an IoU threshold of 0.5 for the teacher network during testing.
\vspace{-1em}
\subsection{Comparison with state-of-the-art methods}
\textbf{Weather Adaptation.} As shown in Table \ref{tab:city}, our method achieves an mAP of 40.2\% on the weather adaptation task, which is the state-of-the-art performance for the source-free DAOD task. Additionally, we find that our method is comparable to or even better than the popular DAOD methods~\cite{chen2018domain,saito2019strong,su2020adapting,he2020domain,deng2021unbiased}, where the source data is accessible.

\noindent
\textbf{Synthesis to Real World Adaptation.} In Table \ref{tab:sim}, we illustrate the performance comparison on the synthesis to real world task. Our method reaches an mAP of 50.1\% with a gain of +4.4\% over the LODS~\cite{li2022source}. Compared with the recent mean-teacher based DAOD method UMT~\cite{deng2021unbiased}, we also achieve a remarkable increase of +7.0\%.

\noindent
\textbf{Camera Configuration Adaptation.} Different camera configurations widely exist in practice, which leads to domain shift. Table \ref{tab:sim} displays the results on the camera configuration adaptation task. Our method improves upon previous state-of-the-art LODS~\cite{li2022source} by +3.9\%.  

\begin{figure}[t]
\centering
\subfigure{
\begin{minipage}[b]{0.23\linewidth}
\includegraphics[width=\linewidth]{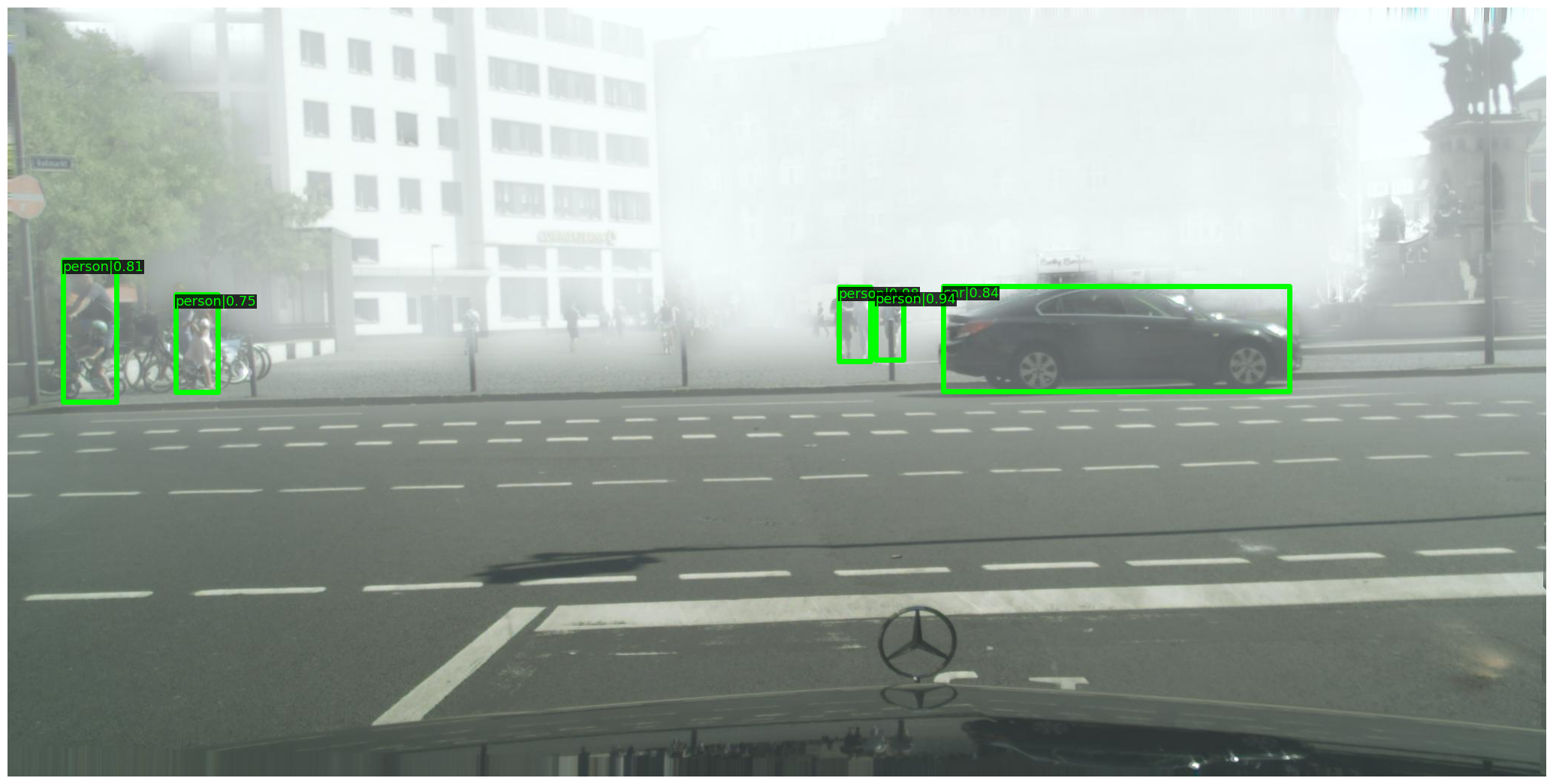}\vspace{1pt}
\includegraphics[width=\linewidth]{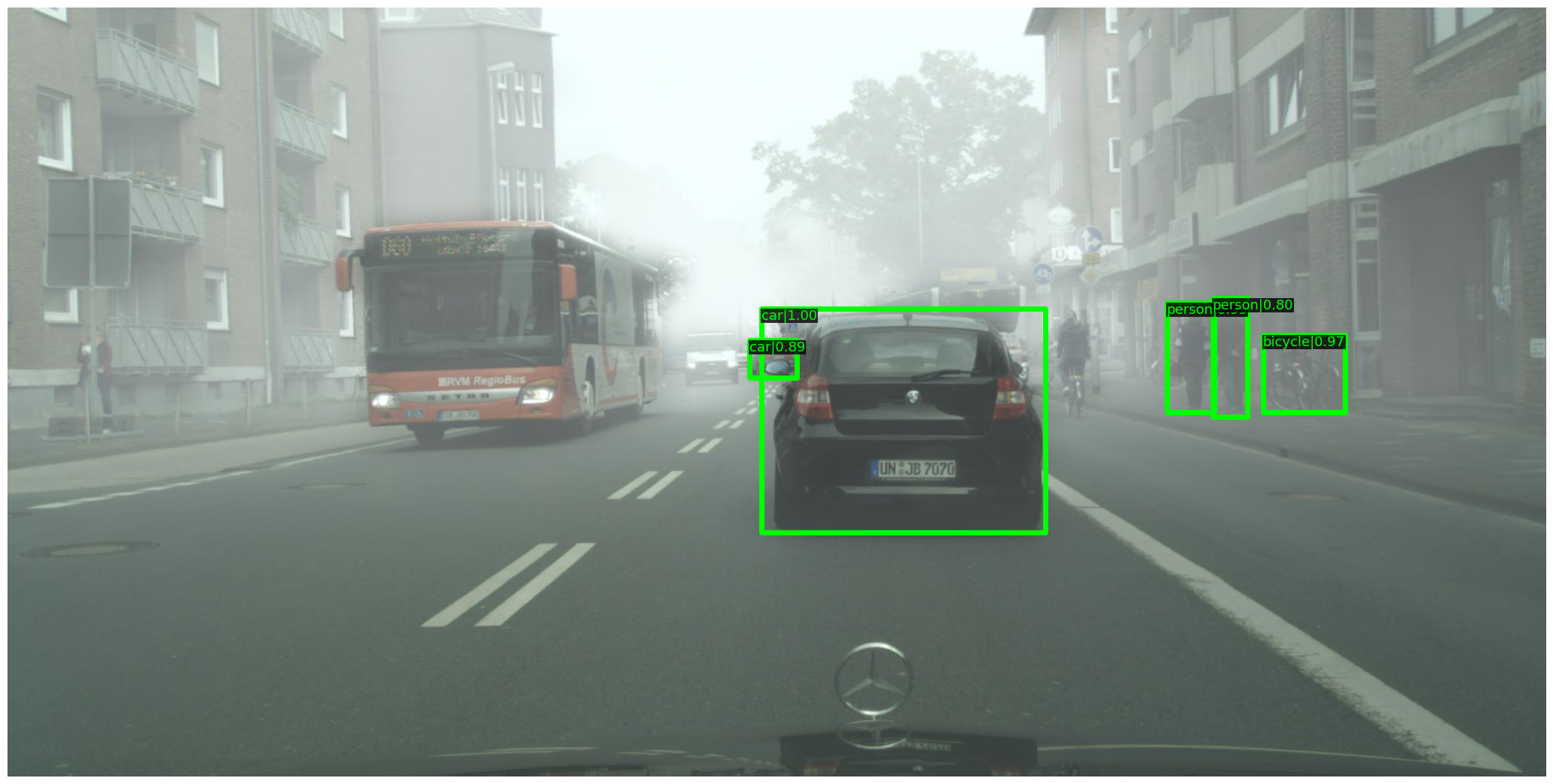}\vspace{1pt}
\includegraphics[width=\linewidth]{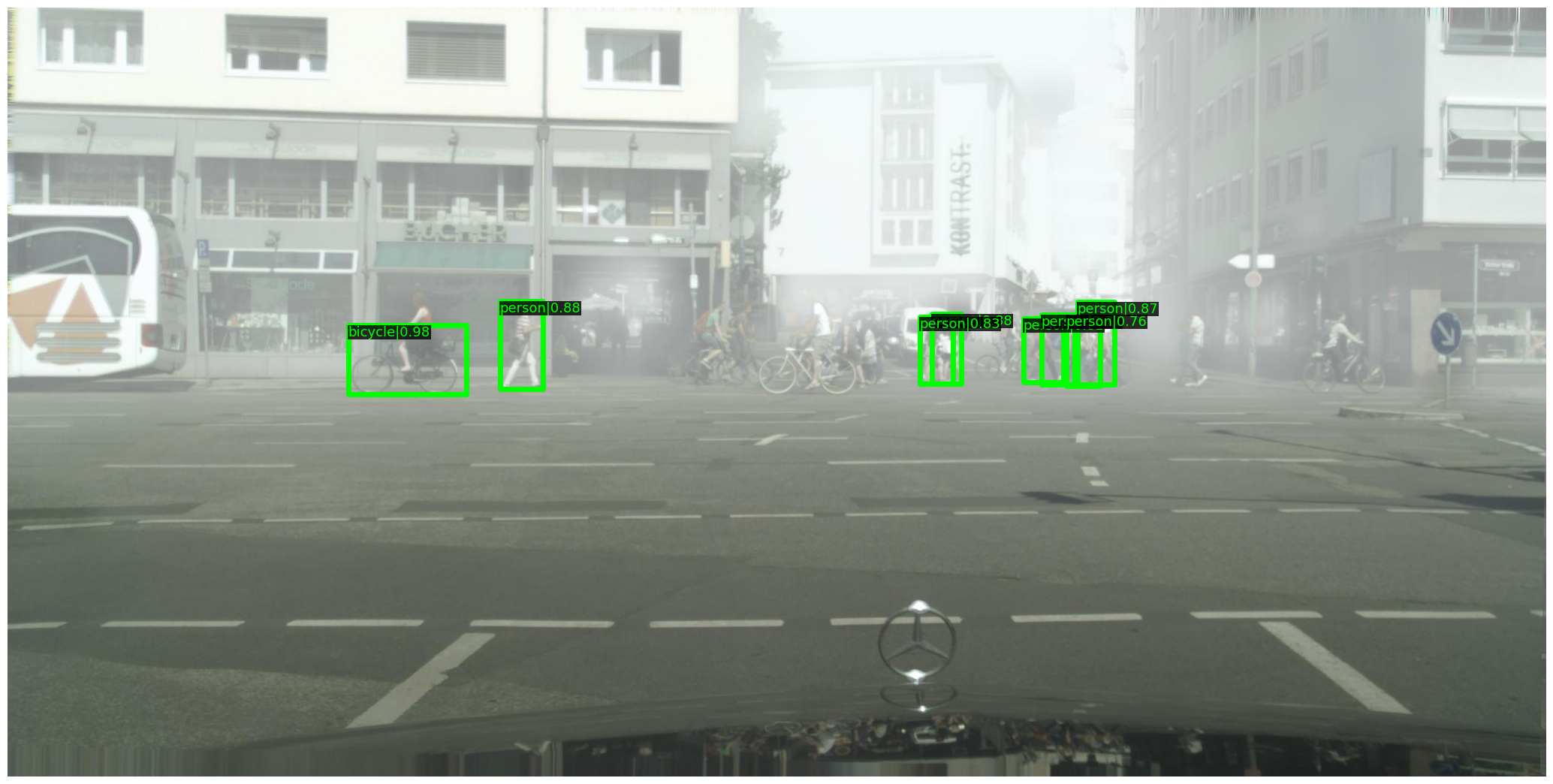}\vspace{-1pt}
\centerline{\scriptsize (a) No DA}
\end{minipage}}
\subfigure{
\begin{minipage}[b]{0.23\linewidth}
\includegraphics[width=\linewidth]{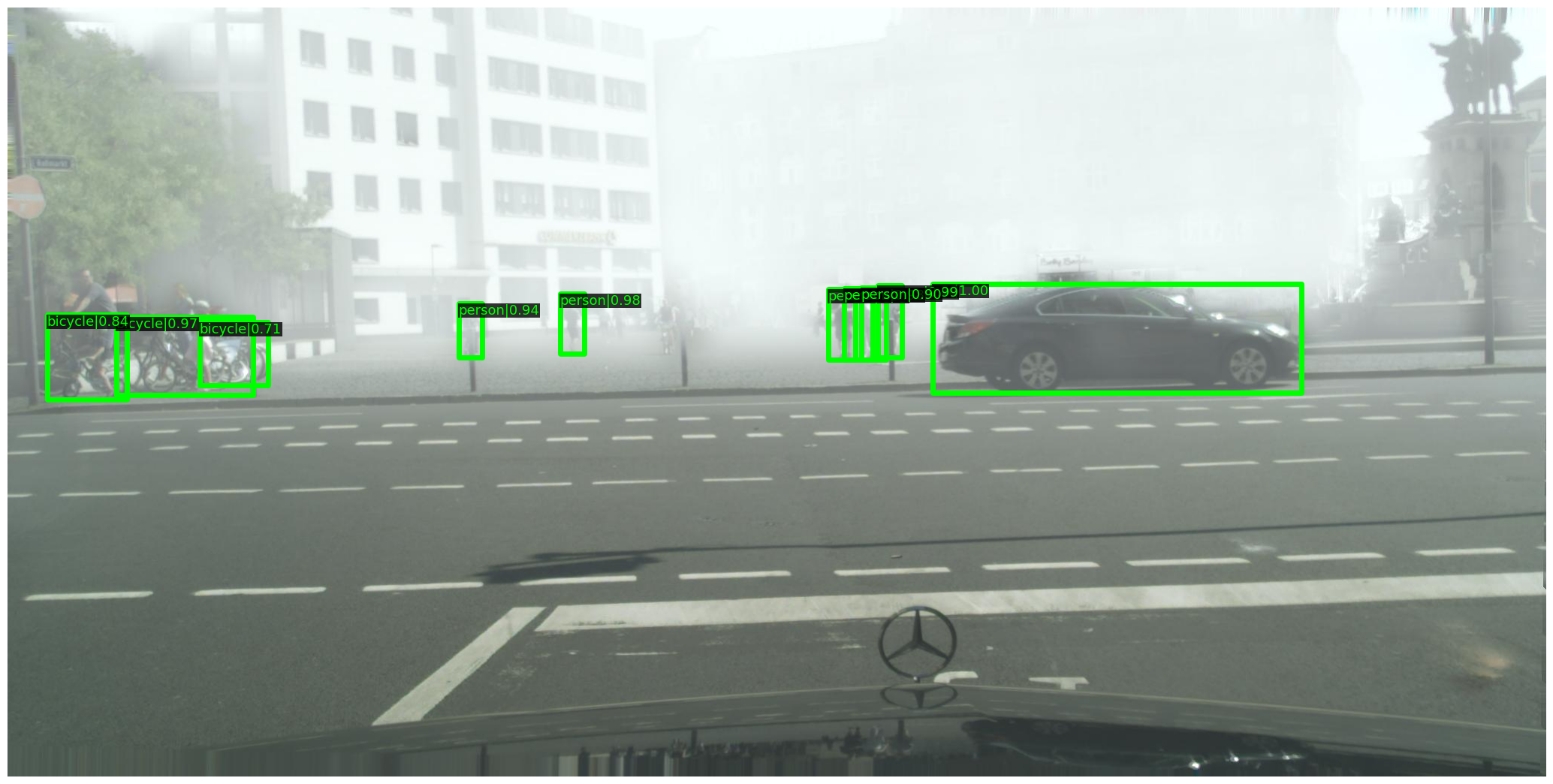}\vspace{1pt}
\includegraphics[width=\linewidth]{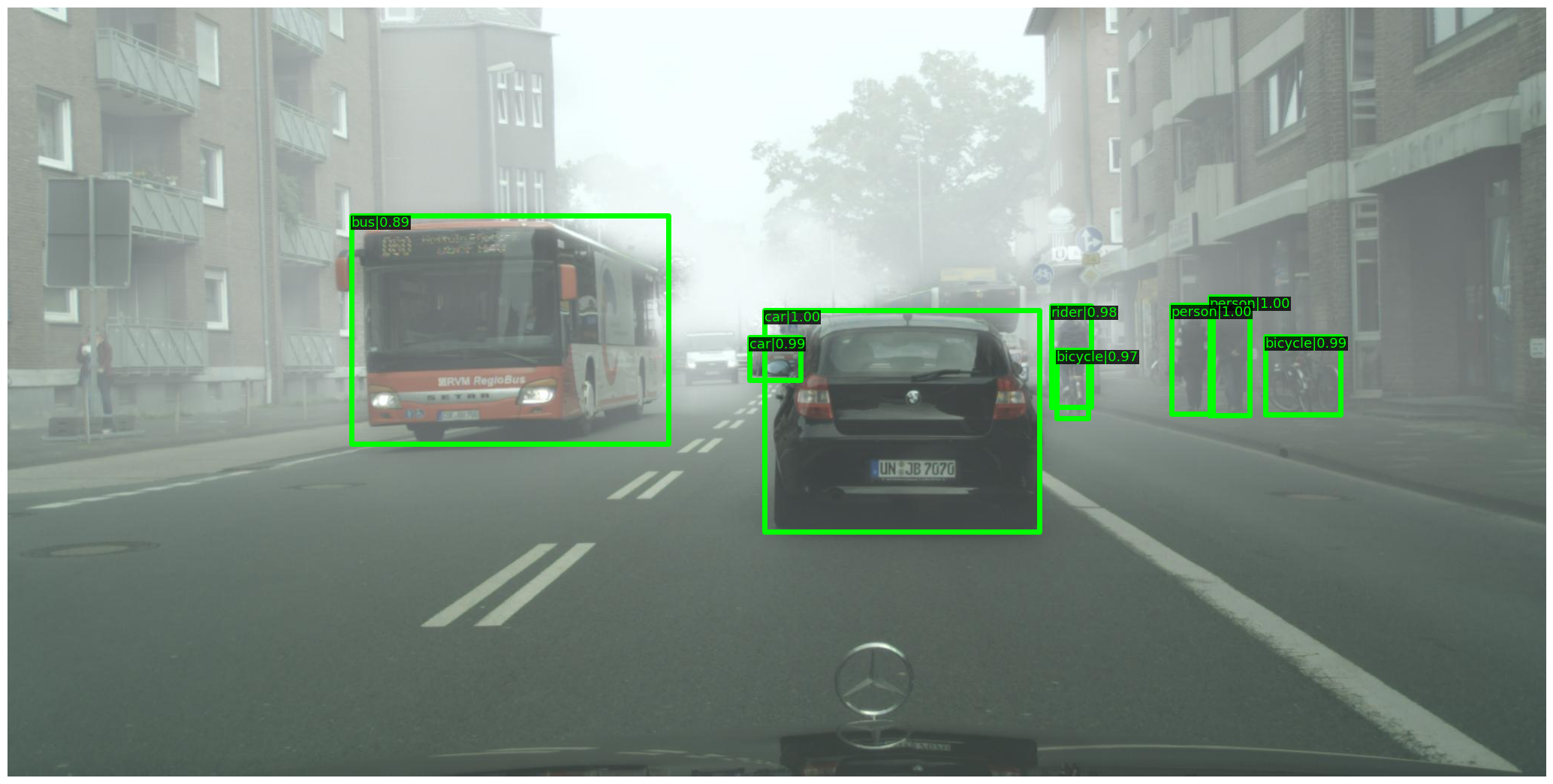}\vspace{1pt}
\includegraphics[width=\linewidth]{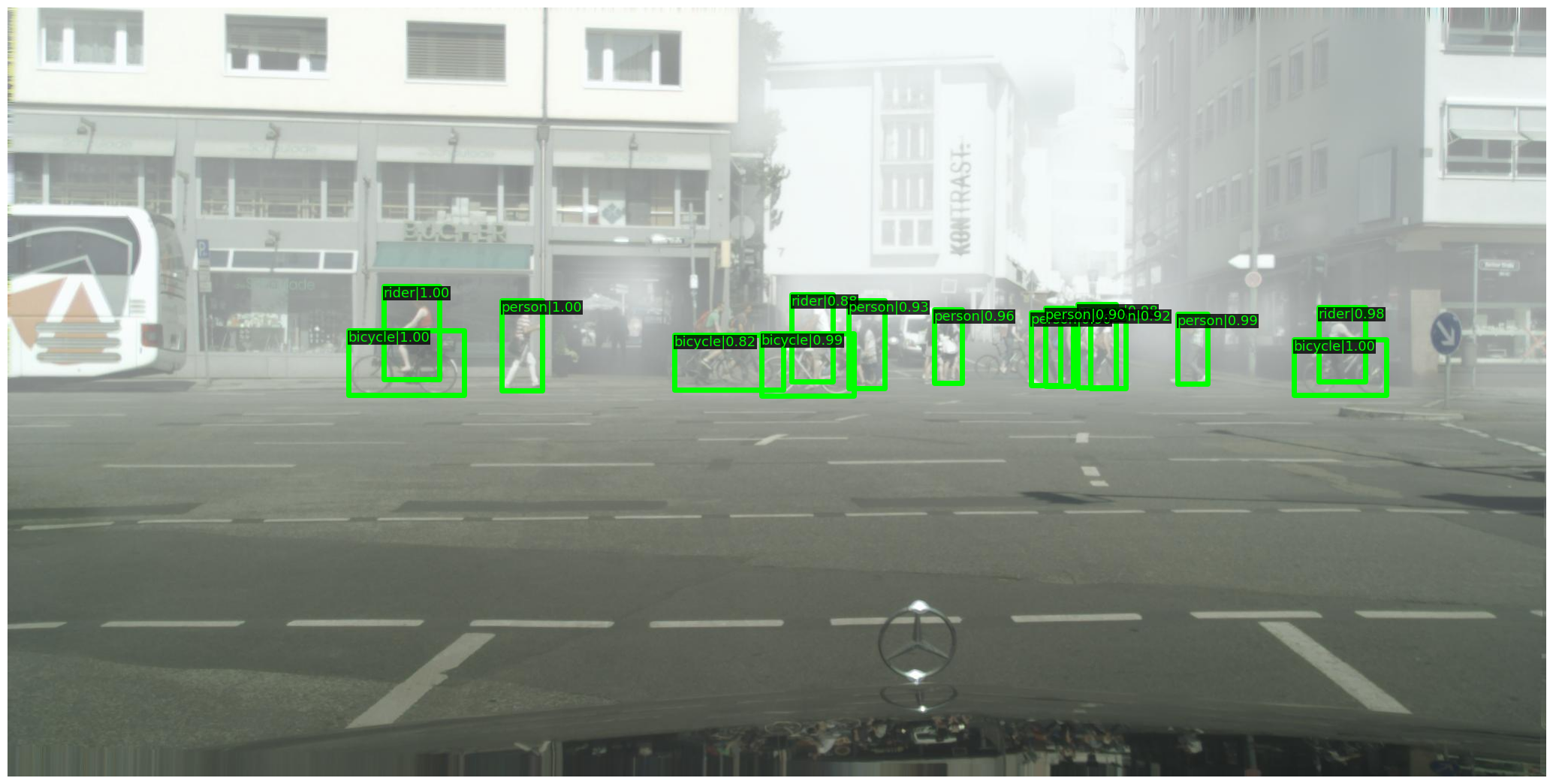}\vspace{-1pt}
\centerline{\scriptsize (b) w/o CATE ($\delta$=0.7)}
\end{minipage}}
\subfigure{
\begin{minipage}[b]{0.23\linewidth}
\includegraphics[width=\linewidth]{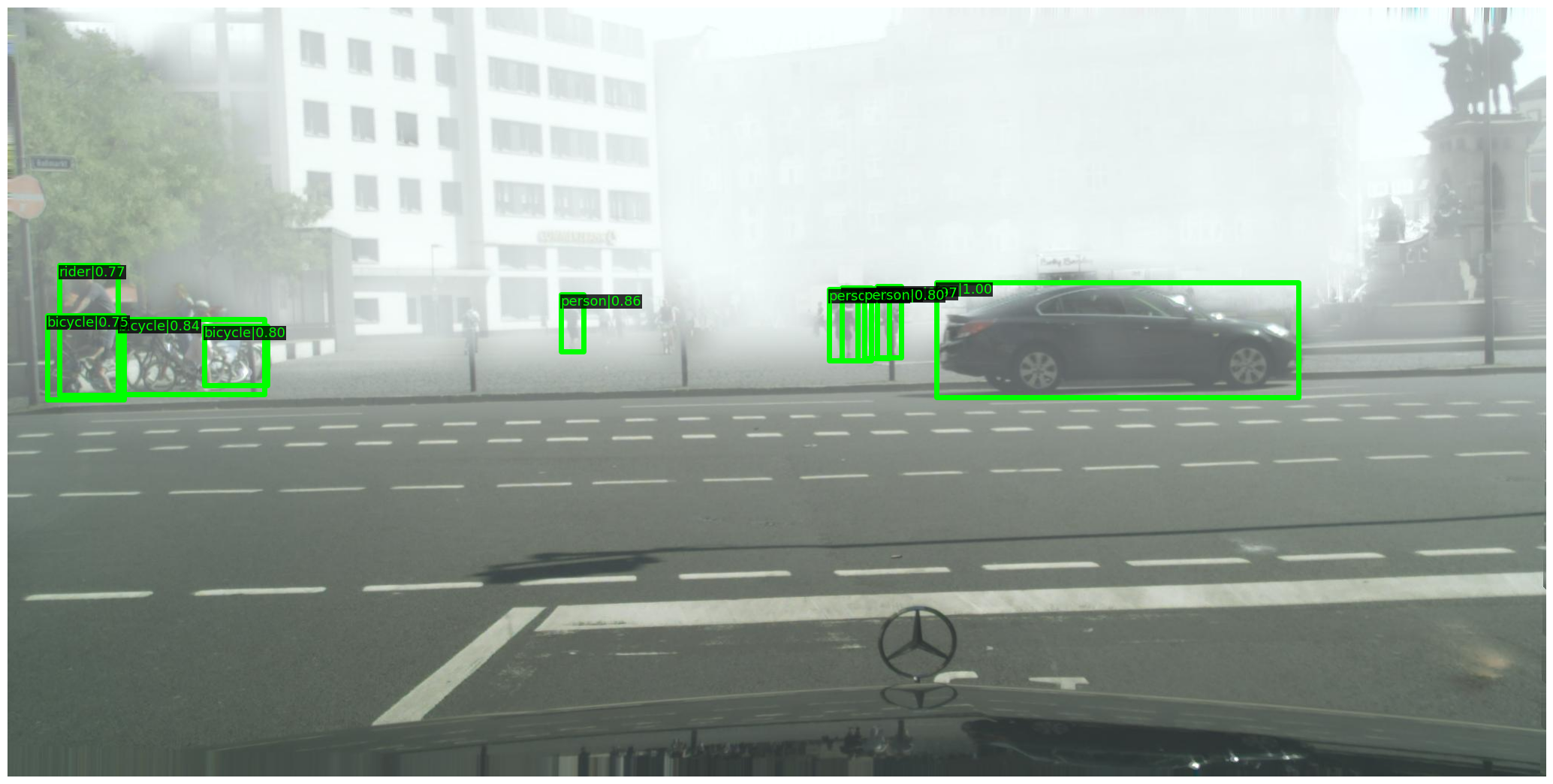}\vspace{1pt}
\includegraphics[width=\linewidth]{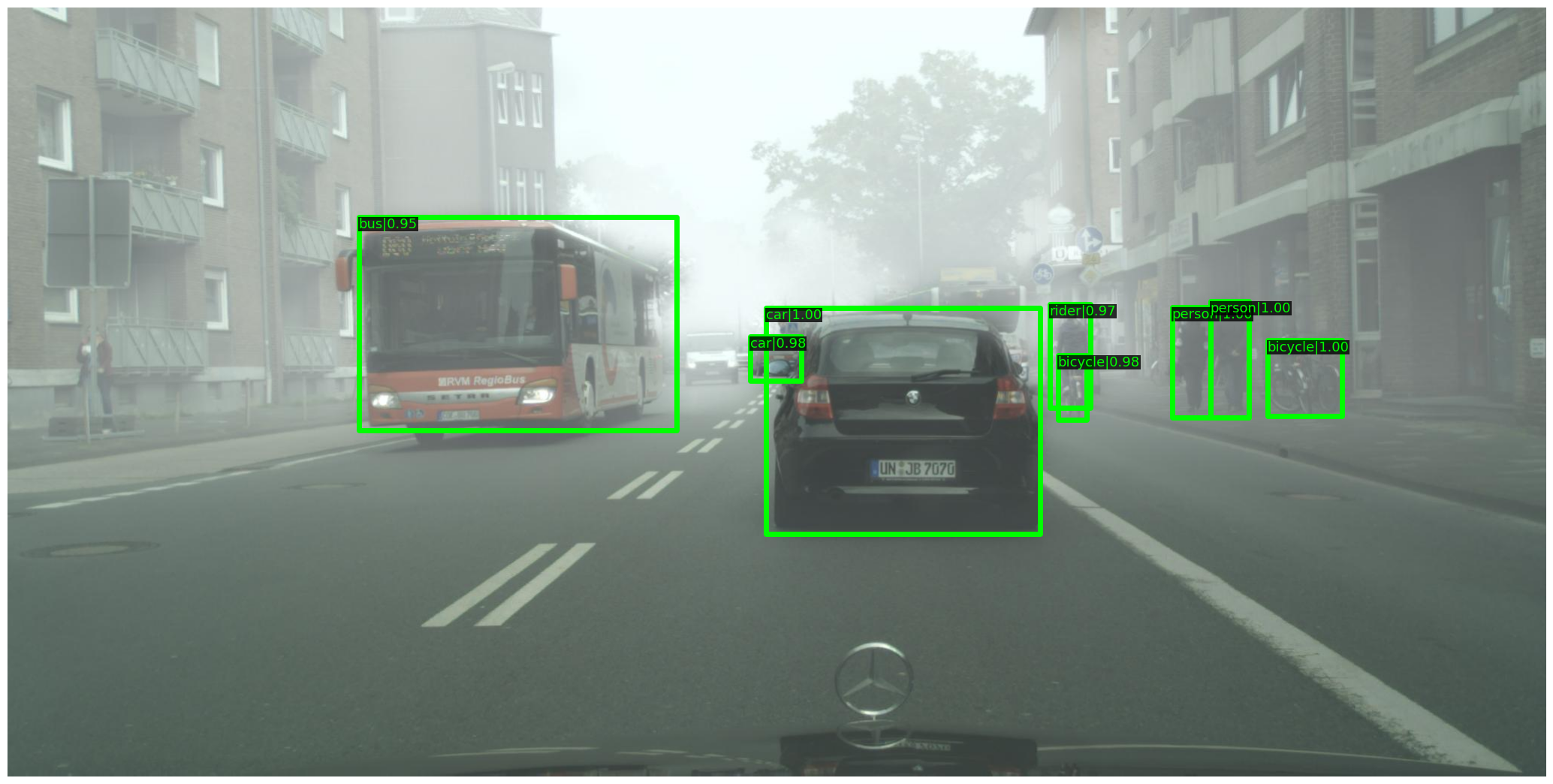}\vspace{1pt}
\includegraphics[width=\linewidth]{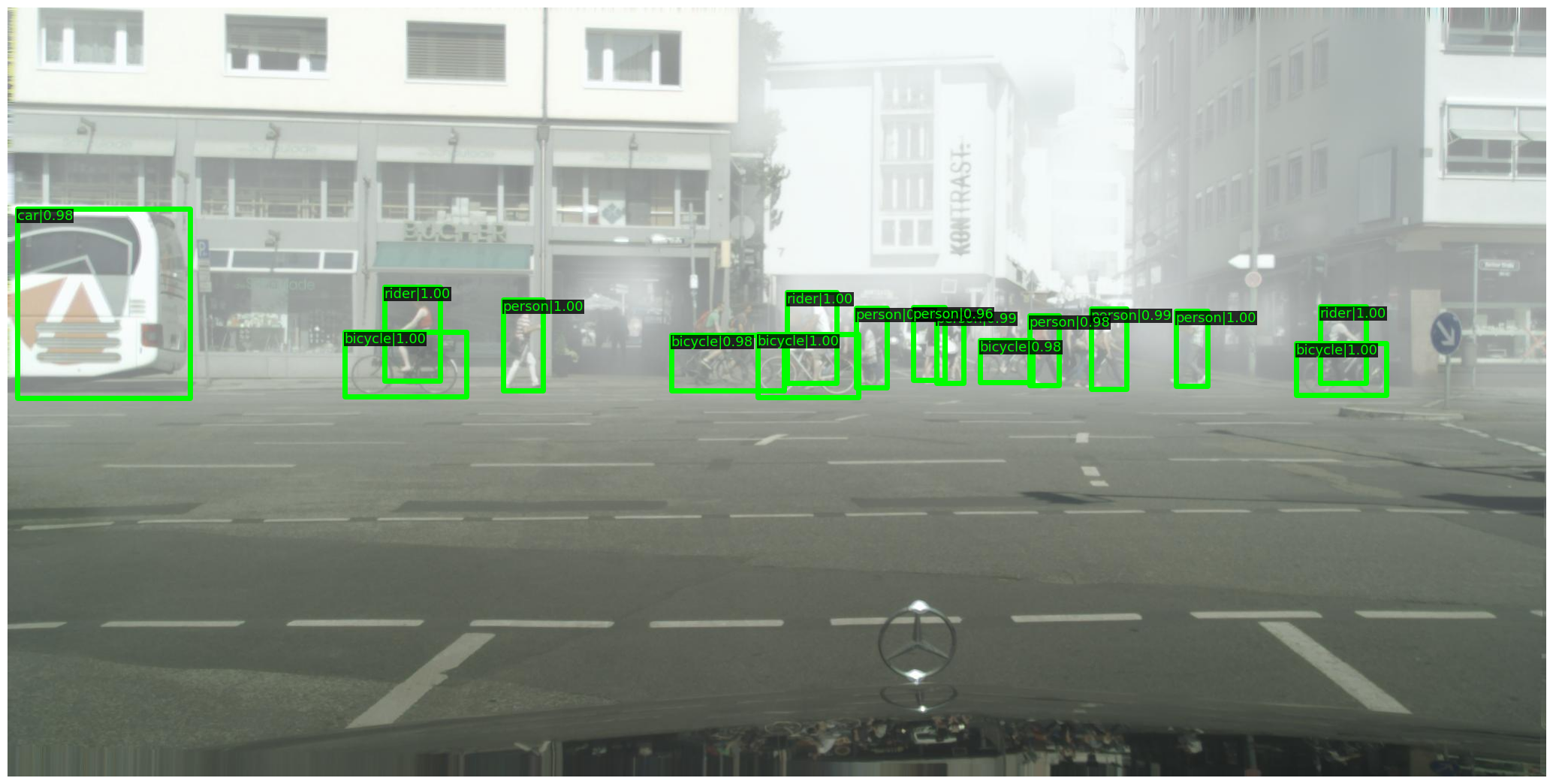}\vspace{-1pt}
\centerline{\scriptsize (c) w/o LPLA}
\end{minipage}}
\subfigure{
\begin{minipage}[b]{0.23\linewidth}
\includegraphics[width=\linewidth]{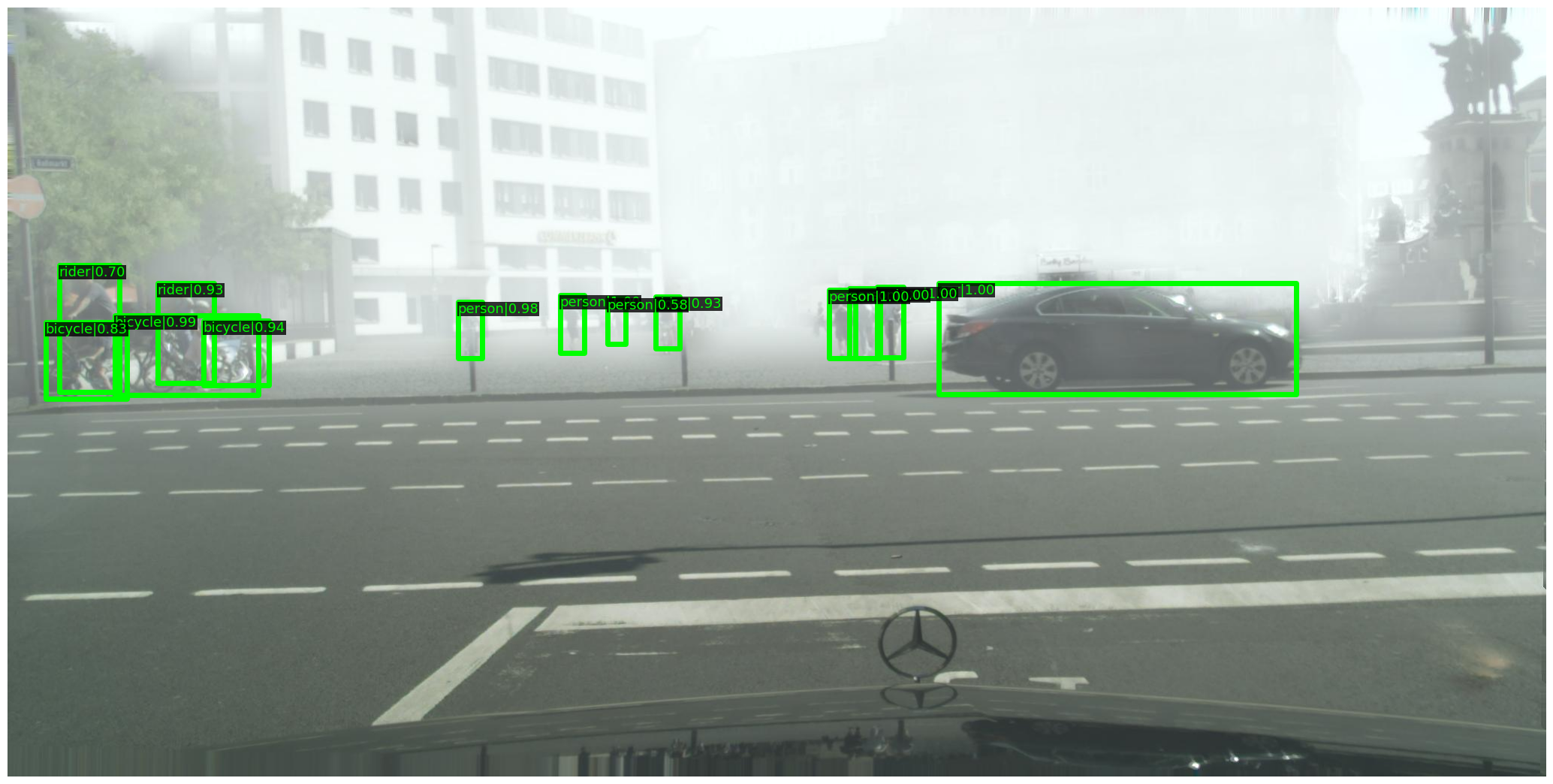}\vspace{1pt}
\includegraphics[width=\linewidth]{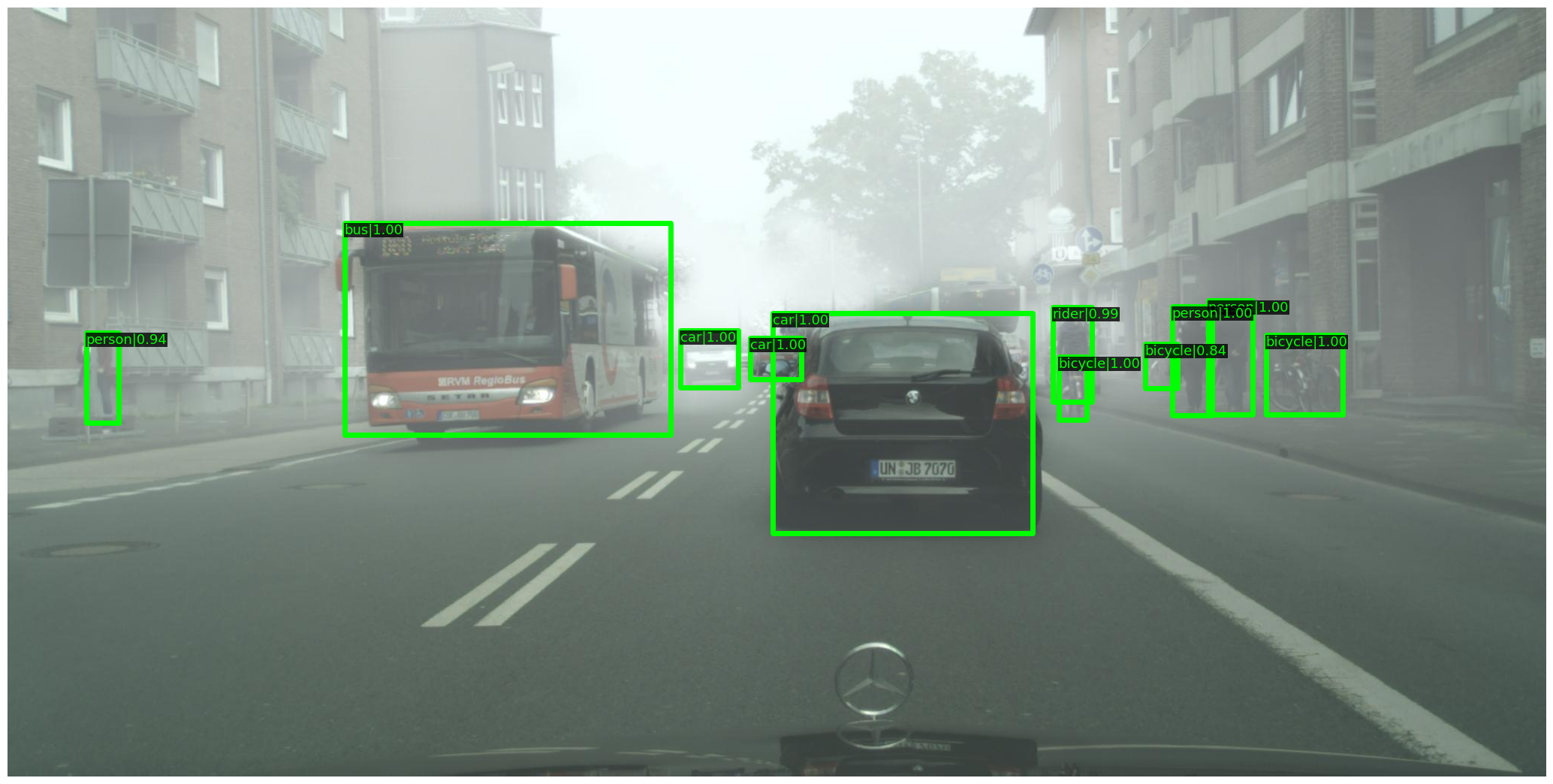}\vspace{1pt}
\includegraphics[width=\linewidth]{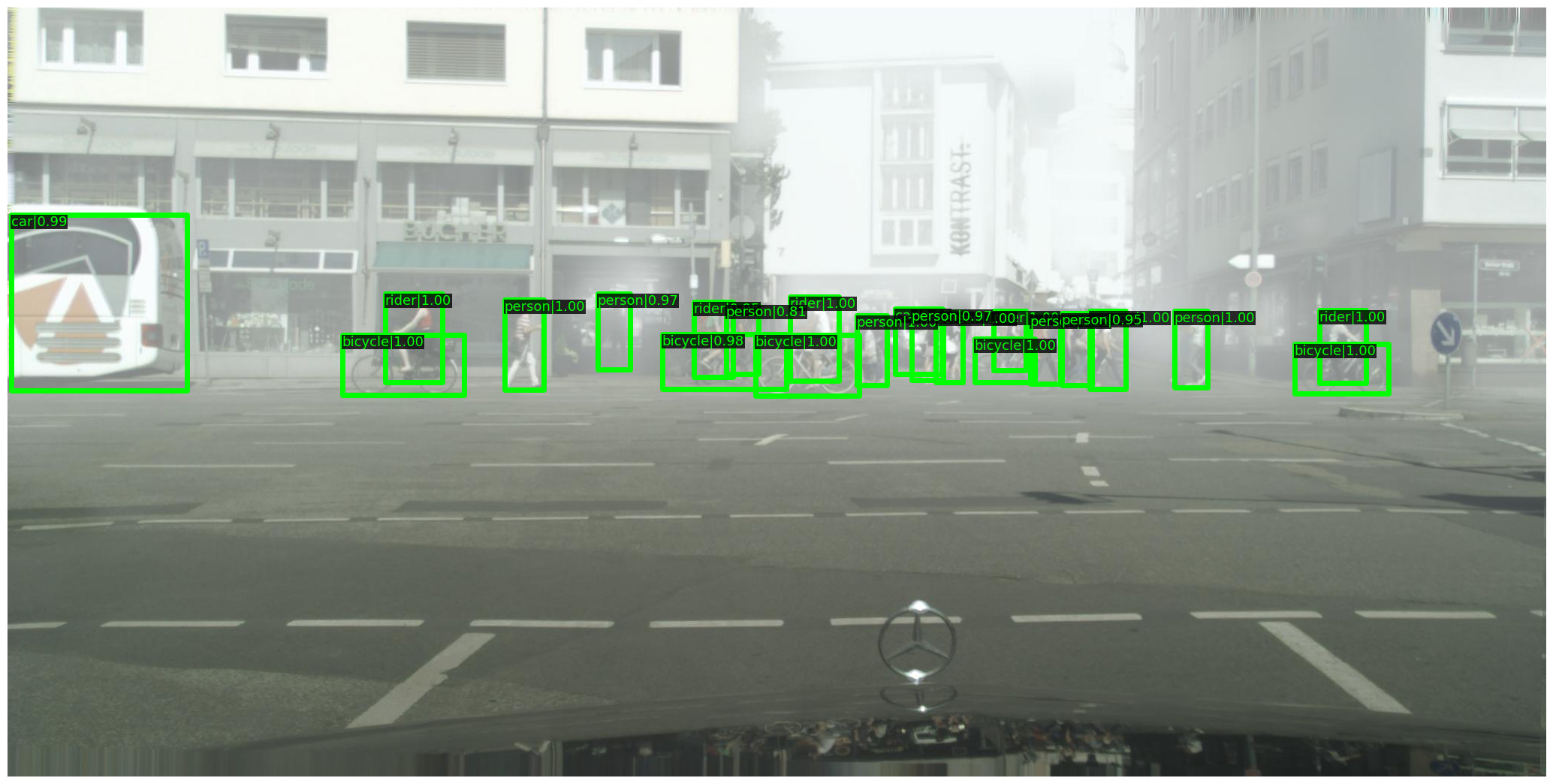}\vspace{-1pt}
\centerline{\scriptsize (d) RPL}
\end{minipage}}
\vspace{-1em}
\caption{Visualizations on the weather adaptation.}
\label{fig3}
\vspace{-1.5em}
\end{figure}
\noindent
\textbf{Real to Art Adaptation.} In this part, we further conduct experiments on adaptation in a huge domain shift from real to clipart-style images, as shown in Table \ref{tab:voc}. Our method achieves 44.0\%, which is inferior to LODS~\cite{li2022source} by 1.2\%, since LODS~\cite{li2022source} employ an extra style transfer module. Compared with recent DAOD methods~\cite{chen2018domain,saito2019strong, xu2020exploring,he2020domain,deng2021unbiased}, our method achieves comparable performance, demonstrating the effectiveness of our method.
\vspace{-1em}
\subsection{Ablation Study}
\textbf{Ablation Study on CATE and LPLA.} Table \ref{tab:ab} shows the effectiveness of the CATE and LPLA module, and it can be observed that they both benefit the detection performance on all four different adaptation tasks. We can see that the performance is sensitive to the selection of the threshold, and our category-aware adaptive threshold is superior to the fixed threshold. Additionally, the effectiveness of LPLA supports that pseudo label assignment based on box regression variance contributes to the generation of reliable pseudo labels. 

\noindent
\textbf{Qualitative Visualization.} In Fig.\ref{fig3}, we show some detection examples of different experimental settings in the adaptation task from Cityscapes to Foggy Cityscapes.  From (b), (c) to (d), we can observe that our method detects small-scale false negatives gradually and true positives get increased, which is consistent with the previous analysis.
\vspace{-1em}
\section{CONCLUSION}
\label{sec:conclusion}
\vspace{-0.5em}
In this paper, we explore the limitations of existing source-free DAOD methods, including category-biased pseudo label generation and inaccuracy localization. To generate unbiased pseudo labels, we propose the category-aware adaptive threshold estimation module. To avoid poor localization, we introduce the localization-aware pseudo label assignment strategy. Experimental results on four different adaptation tasks validate the effectiveness and superiority of our method.
\begin{spacing}{0.89} 

% \begin{spacing}{0.89} 
% \bibliographystyle{ieeetr} 
% \bibliography{refs}
\end{spacing} 

\end{document}